# Towards solving ontological dissonance using network graphs

*Emergent Research Forum (ERF) Paper*

**Maximilian Stäbler**
German Aerospace Center (DLR)
maximilian.staebler@dlr.de

**Frank Köster**
German Aerospace Center (DLR)
frank.koester@dlr.de

**Christoph Schlueter-Langdon**
Drucker School of Business, Claremont Graduate University
Chris.Langdon@cgu.edu

## Abstract

Data Spaces are an emerging concept for the trusted implementation of data-based applications and business models, offering a high degree of flexibility and sovereignty to all stakeholders. As Data Spaces are currently emerging in different domains such as mobility, health or food, semantic interfaces need to be identified and implemented to ensure the technical interoperability of these Data Spaces. This paper consolidates data models from 13 different domains and analyzes the ontological dissonance of these domains. Using a network graph, central data models and ontology attributes are identified, while the semantic heterogeneity of these domains is described qualitatively. The research outlook describes how these results help to connect different Data Spaces across domains.

## Keywords

Data Spaces, Data Space Mesh, Semantic Interoperability, Ontology Dissonance, Semantic Mapping

## Introduction

The growing number of heterogeneous data sources and the increasing scale of information systems due to the transformational development of information and communication technologies are leading to an ever-increasing variety of data (Curry et al. 2022). This enables service innovation based on the collection, processing, and use of data and promotes novel data interoperability concepts such as Data Spaces (Otto et al. 2022). In these Data Spaces, ontologies are used to enable a common understanding of information and data and to make explicit assumptions about the domain so that this data can be used more effectively. Funding projects such as *Gaia-X (*https://gaia-x.eu/) and *Catena-X (*https://catena-x.net/en) aim to create flexible and open IT structures (i.e., Data Spaces). While guaranteeing the full sovereignty of the actors involved, these Data Spaces enable the trusted and transparent use of decentral organized data in accordance with previously defined scopes and goals of use. With numerous Original Equipment Manufacturer, and Tier 1 suppliers participating in these funding projects, these projects have standard-setting power and act as promoters and enforcers of new ontologies (Drees et al. 2022). The usage of domain-, company-, and application-specific ontologies, as well as the combination of these, is important (Curry 2012). This paper is organized as follows: After stating the research questions, the following chapter describes the technical foundations of ontologies, network graphs, and semantic interoperability. Then it is explained how these technologies are used to describe the semantic distance between domains, before presenting the research results obtained so far and classifying current problems of semantic interoperability for Data Spaces. Finally, an outlook on further research approaches is given.

**RQ1:** How can the dissonance between ontologies from different domains be measured and described?
**RQ2:** What opportunities does network graph analysis offer for ontology (semantic) alignment approaches in the context of interoperable Data Spaces?

## Motivation & Theoretical Background

The *Gaia-X* initiative aims to develop digital governance based on European values that can be applied to any existing cloud/edge technology stack to achieve transparency, controllability, portability and interoperability of data and services. *Catena-X* pursues the same goals but focuses on the data-driven value chain within the German automotive industry. To prevent the emergence of monolithic and encapsulated Data Spaces, it is important to consider and prepare for the interconnection of different Data Spaces during the design phase. Interconnected Data Spaces (= Data Space Mesh) are based on Data Spaces and enable the integrated use of data from data sources that are part of different Data Spaces. The resulting cross-sectoral utilization of data can lead to syntactic and semantic conflicts. According to the authors, understanding the cross-domain dissonance of different ontologies, as well as the alignment of these ontologies performed on this basis, is important for semantic interoperability and a basic condition for a successful, automated, and scalable Data Space Mesh.

To capture and provide a qualified description of the dissonance of different ontologies in distinct domains, exchange points (same or similar elements) in different ontologies (i.e., in a network of ontologies) must be identified (Hooi et al. 2014). In research, graphs are used to study networks and capture connections (Otero-Cerdeira et al. 2015). A graph is composed of a node set and an edge set, where nodes represent entities and edges represent the relationship between entities. The nodes and edges form the topology structure of the graph. Besides the graph structure, nodes, edges, and/or the whole graph can be associated with rich information represented as node/edge/graph features (also known as attributes or contents) (Wu et al. 2022). The network perspective provides a set of methods for analyzing the structure of whole entities as well as a variety of theories explaining the patterns observed in these structures. In the further course of this paper, the relevance of an entity to the overall graph is derived based on the degree of the corresponding node (number of edges to and from the node). Moreover, for semantic harmonization of different ontologies, the similarity of individual sub-graphs (an ontology-graph within the overall network graph) can be used as a similarity measure (Wu et al. 2022).

The literature describes the potential of ontology alignment for interoperability of data from heterogeneous sources (Otero-Cerdeira et al. 2015). In this context, ontologies form a layer of abstraction over data, taxonomies, or database schemas and provide richer semantics through concept mapping. Ontology alignment is a complex process that helps to reduce the semantic gap between different overlapping representations across domains (Hooi et al. 2014).

Alignment helps to establish a mapping between entities that semantically belong to different ontologies. Most ontology alignment approaches use elementary matching techniques (e.g., string-based methods, linguistic methods, etc.). These techniques map elements by analyzing entities in isolation and ignoring their relationships to other entities (father/son, brother, etc.). Bypassing the latter aspects, determining the semantics of an entity is often difficult. Therefore, the structural information of an ontology (e.g., by mapping it in a network graph) plays an important role in ontology mapping (Ouali et al. 2019). To achieve semantic interoperability of Data Spaces (each with a high degree of self-determination), the ontologies used are augmented with data about the data (metadata). The inconsistent understanding that arises from different interpretations of the ontology parts when describing the data cannot be prevented in the real world (Drees). For example, if one wants to connect existing data sources from a company to an existing Data Space or if one is forced to use a different ontology to represent domain-specific features due to large domain differences, a complex and time-consuming manual ontology alignment is required.

Consequently, this paper is not concerned with an academic consideration of a semantic landscape, but with a practical problem and relevance in the real world. This observation was made by the authors.

The problem has been identified, but the treatment and solution of semantic heterogeneity in the context of a Data Space mesh is not part of this paper. This paper focuses on the qualified description of the dissonance of ontologies from different domains. Further steps are addressed in the research outlook.

## Technical Analysis

A cross-domain set of data models for interoperable and reproducible smart solutions is provided by the Smart Data Models Initiative (https://smartdatamodels.org/). 122 individual contributors, from 80 organizations were able to collect more than 1000 *Data Models*, divided into 13 *Domains* such as *SmartCities, Health* or *Agriculture & Food (AgriFood)*. A special component is the *CrossSector* Domain, which bundles *DataModels* from different *Domains*. The *SmartDataModels* (SDM) adopt proven models (RDF Vocabularies) from open and adopted standards such as *DCAT-AP (https://www.w3.org/TR/vocab-dcat-2/)* or *CPSV-AP (https://semiceu.github.io/CPSV-AP/releases/3.1.0/)*. Overall, the SDM form a selection of domain-specific ontologies and are therefore evaluated as a suitable data basis for evaluating and qualifying the problems of semantic interoperability for Data Spaces.

The *DataModels* consist of at least one *Type*, which in turn has at least one *Attribute* as a property. The SDM provided on *Github* were crawled for this paper and transferred into a MongoDB database with the described structure by a manual Python implementation. In the resulting network graph, the *Domains*, the *DataModels*, *Types* and the *Attributes* are represented as Nodes. Table 1 gives an overview of the hierarchy of the different node types and components of the SDM:

| **NodeType** | *Domain* | *DataModel* | *Type* | *Attribute* |
|---|---|---|---|---|
| **Description** | Activity sector | Technical representation of an entity related to an activity within a domain | Subgroup within a DataModel | Property of a Type |
| **Example** | *SmartCities* | *UrbanMobility* | *ArrivalEstimation* | *[dataProvider, hasTrip]* |

**Table 1. Node type overview**

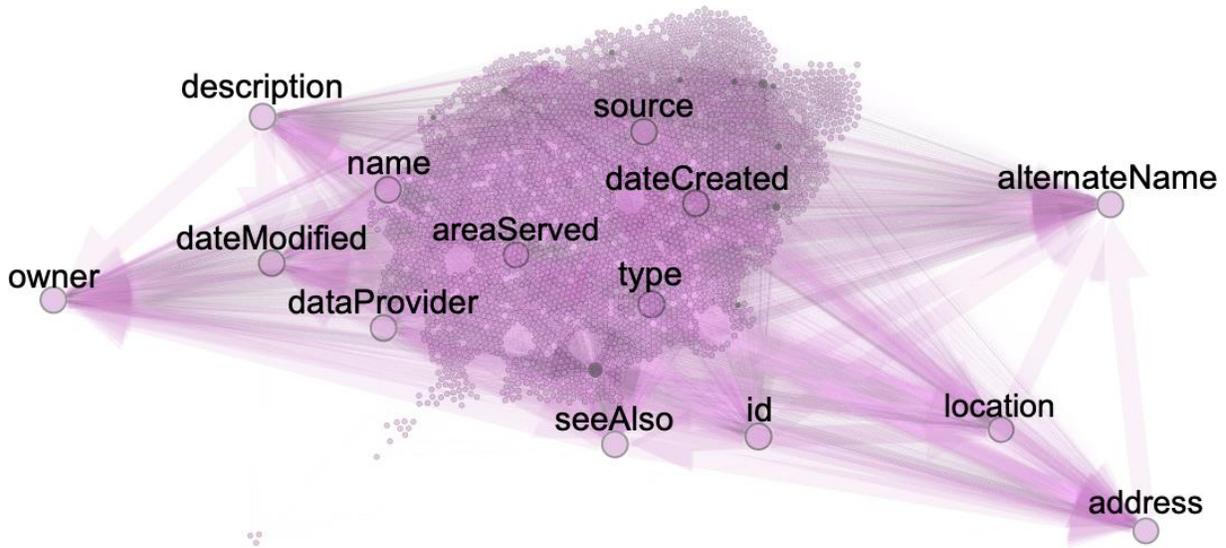

**Figure 1. Sub-Graph showing all *Attributes* of the SDM including the labels of the 14 *Attributes* with the highest degree.**

For each *Attribute* in each *DataModel*, edges are drawn to the other *Attributes* within the same *Type*. Also, one edge to both, the parent *DataModel* and the *Domain*. For the given examples in Table 1, an edge is drawn between *dataProvider* and *hasTrip* as well as an edge to *ArrivalEstimation* and *SmartCities* for each of both *Attributes*. The resulting graph has a total of 212871 edges and 3630 nodes which represent the connections and relationships of 13 domains, 59 data models, 62 types and 3496 attributes. A detailed overview of the *Attributes*-graph can be seen in Figure 1 and the *Domains* considered can be seen in the heatmap in Figure 2. It is important to note here that the *DataModels* are not equally divided between the domains.

The graph structure offers possibilities to examine connections between individual nodes and enables conclusions to be drawn about the cross-domain use (therefore importance) of individual *Attributes*. In addition, it can be shown how similar different *Domains* are regarding the shared usage of certain

*Attributes*. In Figure 1, it can be seen that a few *Attributes* contrast strongly with all other *Attributes* due to their central position within the graph and high degree (size of node). Centrality metrics such as *Degree Centrality* and *Betweeness Centrality* are recognized methods to identify central (important) nodes within a graph (Zhang and Luo 2017). Using these two metrics, 14 attributes were identified that are used across domains in many *DataModels*. The black dots (nodes) represent the 13 domains. The positioning of the domains does not allow clear interpretations.

Figure 2 shows the use of common data models and attributes (vocabularies) at the domain level. The 13 unique domains are shown vertically and horizontally. The field where a horizontal and vertical column meet contains the number of shared *DataModels*. The diagonal of the heatmap is filled with 0. The number line above the heatmap shows the range of the distinct counts.

One can see that there are domains that have many more common data models than others. For example, *SmartCities* and the *SmartEnvironment* domain have much more overlap than, like *SmartCities* and *SmartRobotics*. Likewise, it seems to generally manifest that there are domains that tend to have more overlap with several other *Domains* and *Domains* that tend to be more isolated from other *Domains* and ergo tend to use more specialized ontologies (vocabularies). An example of the latter would be *SmartRobotics*. The *CrossSector* domain contains a particularly large number of *DataModels* that are also available in the *SmartCities* and *SmartEnergy* domains. One explanation for this could be the size and scope of these two *Domains*, as well as the focus of the SDM on the *SmartCities* domain. It can be equally noted that the majority of *Domains* tend to use specific (i.e. not shared with others) *DataModels*.

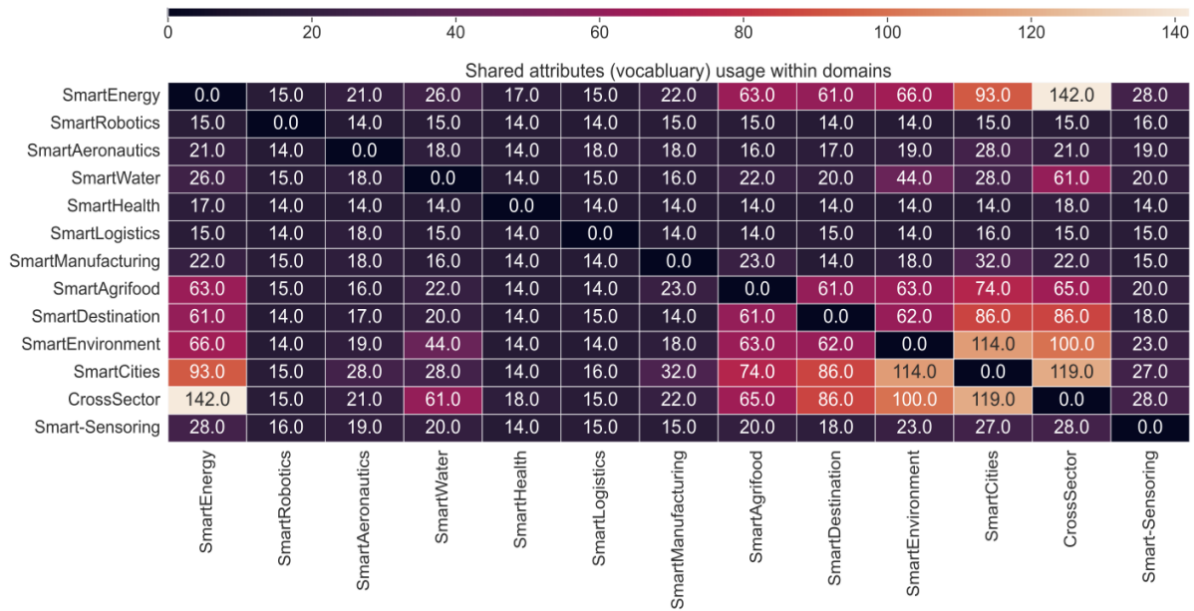

**Figure 2. Overview of the *Attributes* connections between all 13 *Domains*. The color bar on top shows the connection count between two *Domains*.**

## Discussion of intermediate Results

The present work has shown how dissonance, both between ontologies within a domain and between ontologies in different domains, can be described in a qualified way (RQ1). Likewise, it has been shown what possibilities arise from describing the semantics in different domains in a graph with respect to the alignment of ontologies and therefore for interoperable Data Spaces. Graphs can not only identify important and similar attributes in different ontologies and domains, but there is also the possibility to extend these graphs with meta-data (RQ2). The analysis of the cross-domain graph derived from the SDM shows that semantic models of different domains are similar in a few attributes, but mostly use domain-specific attributes. Although the domains in the created dataset contain a different number of data models, it can still be observed that there are domains with specific vocabularies and domains in which the dissonance with other domains is not strong (i.e., easier to harmonize with other ontologies). Since cross-organizational, cross-domain, and cross-country Data Spaces do not use structurally identical data models of one initiative as in the present analysis, but ontologies and semantic models of different authors and different contexts, it can be concluded that automated ontology alignment approaches are important for the successful construction of interoperable Data Spaces (Data Space Mesh). There is not only a need for ontology alignment approaches at the structure level, but also at the element level.

## Next steps and Outlook

To solve the complex information sharing and contextual understanding of ontologies in a Data Space Mesh, further research will investigate the semantic mapping approach described by (Kotis et al. 2006). Also, Graph Matching Learning (GML), according to *Wu et al.* (Wu et al. 2022) will be considered for

(1) optimal node-to-node correspondence between nodes of a pair of graphs and

(2) the graph similarity problem which computes a similarity metric between two graphs.

To the best of our knowledge, a description of specific semantic model requirements for building a Data Space Mesh and to realize an automated process for semantic interoperability has not yet been published. This process is necessary to ensure the scalability of a Data Space Mesh and to make the solution resilient.